\documentclass[sigconf]{acmart}

\copyrightyear{2022} 
\acmYear{2022} 
\setcopyright{acmcopyright}\acmConference[WWW '22]{Proceedings of the ACM Web Conference 2022}{April 25--29, 2022}{Virtual Event, Lyon, France}
\acmBooktitle{Proceedings of the ACM Web Conference 2022 (WWW '22), April 25--29, 2022, Virtual Event, Lyon, France}
\acmPrice{15.00}
\acmDOI{10.1145/3485447.3512034}
\acmISBN{978-1-4503-9096-5/22/04}

\usepackage{url}

\usepackage{multirow}
\usepackage{balance}
\usepackage{amsfonts}
\usepackage{amsmath}
\usepackage{amsthm}
\usepackage{graphicx}
\usepackage{microtype}
\usepackage{url}
\usepackage{enumitem}
\usepackage{wrapfig,lipsum}
\usepackage{multirow}
\usepackage{makecell}
\usepackage{wrapfig}
\usepackage{tabularx}
\usepackage{subfigure}
\usepackage[noend,ruled]{algorithm2e}
\usepackage{bbm}
\usepackage{algpseudocode}
\usepackage{etoolbox}
\usepackage{subfigure}
\usepackage{float}
\usepackage{booktabs}
\usepackage{caption}
\usepackage{setspace}
\usepackage[makeroom]{cancel}

\newcommand{\bs}[1]{\boldsymbol{#1}}

\renewcommand{\textrightarrow}{$\rightarrow$}

\newcommand{\ie}{\emph{i.e.}}
\newcommand{\eg}{\emph{e.g.}}
\newcommand{\model}{\mbox{\sf TopClus}\xspace}
\newcommand{\wrong}[1]{\underline{\textit{#1}}}

\AtBeginDocument{%
  \providecommand\BibTeX{{%
    \normalfont B\kern-0.5em{\scshape i\kern-0.25em b}\kern-0.8em\TeX}}}

\author{Yu Meng, Yunyi Zhang, Jiaxin Huang, Yu Zhang, Jiawei Han}
\affiliation{
\institution{Department of Computer Science, University of Illinois at Urbana-Champaign, IL, USA} 
\country{}
\institution{\{yumeng5, yzhan238, jiaxinh3, yuz9, hanj\}@illinois.edu}
}

\renewcommand{\shortauthors}{Meng et al.}

\begin{document}
\fancyhead{}
\title{Topic Discovery via Latent Space Clustering of Pretrained Language Model Representations}



\begin{spacing}{0.96}

\begin{abstract}
Topic models have been the prominent tools for automatic topic discovery from text corpora. Despite their effectiveness, topic models suffer from several limitations including the inability of modeling word ordering information in documents, the difficulty of incorporating external linguistic knowledge, and the lack of both accurate and efficient inference methods for approximating the intractable posterior. 
Recently, pretrained language models (PLMs) have brought astonishing performance improvements to a wide variety of tasks due to their superior representations of text.
Interestingly, there have not been standard approaches to deploy PLMs for topic discovery as better alternatives to topic models.
In this paper, we begin by analyzing the challenges of using PLM representations for topic discovery, and then propose a joint latent space learning and clustering framework built upon PLM embeddings.
In the latent space, topic-word and document-topic distributions are jointly modeled so that the discovered topics can be interpreted by coherent and distinctive terms and meanwhile serve as meaningful summaries of the documents.
Our model effectively leverages the strong representation power and superb linguistic features brought by PLMs for topic discovery, and is conceptually simpler than topic models.
On two benchmark datasets in different domains, our model generates significantly more coherent and diverse topics than strong topic models, and offers better topic-wise document representations, based on both automatic and human evaluations.\footnote{Code and data can be found at \url{https://github.com/yumeng5/TopClus}.}
\end{abstract}

\ccsdesc[500]{Information systems~Clustering}
\ccsdesc[500]{Information systems~Document topic models}
\ccsdesc[500]{Computing methodologies~Natural language processing}

\keywords{Topic Discovery, Pretrained Language Models, Clustering}


\maketitle

\section{Introduction}

Automatically discovering coherent and meaningful topics from text corpora is intuitively appealing for web-scale content analyses, as it facilitates many web applications including document analysis~\cite{blei2007correlated}, text summarization~\cite{wang2009multi} and ad-hoc information retrieval~\cite{wang2007topical}.
Decades of research efforts have been dedicated to the development of such algorithms, among which topic models~\cite{Blei2003LatentDA,hofmann2004latent} are the most prominent methods. 
The success of topic models can be largely credited to their proposed generative process: By maximizing the likelihood of a probabilistic process that models how documents are generated conditioned on the hidden topics, topic models are able to uncover the latent topic structures in the corpus.

Despite the success of topic models, the generative process incurs several notable limitations: 
(1) The ``bag-of-words'' generative assumption completely ignores word ordering information in text, which is essential for defining word meanings~\cite{firth57synopsis}. 
(2) The generative process cannot leverage external knowledge to learn word semantics, which may miss important topic-indicating words if they are not sufficiently reflected by the co-occurrence statistics of the given corpus, as is likely the case for small-scale/short-text corpora. 
(3) The generative process induces an intractable posterior that requires approximation algorithms like Monte Carlo simulation~\cite{neal1993probabilistic} or variational inference~\cite{attias2000variational}. Unfortunately, there is always a trade-off between accuracy and efficiency with these approximations since they can only be asymptotically exact~\cite{salimans2015markov}. 
Later variants of topic models attempt to overcome some of these limitations by either replacing the analytic approximation of the posterior with deep neural networks~\cite{miao2016neural,srivastava2017autoencoding,wang2020neural} to improve the effectiveness and efficiency of the inference process, or incorporating word embeddings~\cite{das2015gaussian,dieng2020topic,nguyen2015improving} to make up for the representation deficiency of the ``bag-of-words'' generative assumption. 
Nevertheless, without fundamental changes of the topic modeling framework, none of these approaches address the limitations of topic models all at once.

Along another line of text representation learning research, text embeddings have achieved enormous success in a wide spectrum of downstream tasks. The effectiveness of text embeddings stems from the learning of distributed representations of words and documents from contexts. 
Early models like Word2Vec~\cite{Mikolov2013DistributedRO} learn context-free word semantics based on a local context window of the center word. 
Recently, pretrained language models (PLMs) like BERT~\cite{devlin2019bert}, RoBERTa~\cite{liu2019roberta} and XLNet~\cite{yang2019xlnet} have revolutionized text processing via learning contextualized word embeddings. They employ Transformer~\cite{vaswani2017attention} as the backbone architecture for capturing the long-range, high-order semantic dependency in text sequences, yielding superior representations to previous context-free embeddings. Since these PLMs are pretrained on large-scale text corpora like Wikipedia, they carry superb linguistic features that can be generalized to almost any text-related applications.

Motivated by the strong representation power of the contextualized embeddings that accurately capture word semantics, a few recent studies have attempted to utilize PLMs for topic discovery. 
Sia et al.~\cite{sia2020tired} directly cluster averaged BERT word embeddings to obtain word clusters as topics. The resulting topic quality relies significantly on heuristic tricks like frequency-based weighting/re-ranking and barely reaches the performance of LDA, the most basic topic model. 
Instead of clustering word embeddings, BERTopic~\cite{grootendorst2020bertopic} clusters document embeddings and then uses TF-IDF metrics to extract representative terms from each notable document cluster as topics.
However, as the document embeddings in BERTopic are obtained from Sentence-BERT~\cite{reimers2019sentence}, which is trained on natural language inference datasets with manually annotated sentence labels, the performance of BERTopic may suffer from domain shift when the target corpus is semantically different from the Sentence-BERT training set, and when manually annotated labels for re-training the sentence embeddings are absent. 
Moreover, BERTopic constructs topics via TF-IDF metrics and fails to take advantage of the distributed representations of PLMs, which are known to better capture word semantics than frequency-based statistics. 

In this work, we study topic discovery with PLM embeddings as a potential alternative to topic models. 
We first analyze the challenges of directly operating on the PLM embedding space by investigating its structure. 
Motivated by the challenges, we propose \model, a joint latent space learning and clustering approach that derives a \emph{lower-dimensional}, \emph{spherical} latent embedding space with topic structures. 
Such latent space mitigates the ``curse of dimensionality'' issue and uses angular similarity to model semantic correlations among words, documents and topics, thus is better suited for clustering than the high-dimensional Euclidean embedding space of PLMs.
Unlike traditional clustering algorithms that work with fixed data representations, \model jointly adjusts the latent space representations and performs clustering. 
Topic-word and document-topic distributions are jointly modeled in the latent space to derive topics that (1) are interpretable by coherent and distinctive words and (2) serve as meaningful summaries of documents.

\model enjoys the following advantages over topic models: 
(1) \model works with PLM contextualized embeddings obtained by modeling the entire text sequences with positional information, which are expected to provide better representations than the ``bag-of-words'' assumption of topic models.
(2) \model employs PLMs to bring in general linguistic knowledge which helps generate more accurate and stable word representations on the target corpus than training topic models from scratch on it.
(3) The training algorithm of \model does not involve any probabilistic approximations, and is computationally and conceptually simpler than variational inference in topic models.
With these advantageous properties, \model simultaneously addresses the major limitations of topic models.

Our contributions are summarized as follows:

\begin{enumerate}[leftmargin=*]
\item We explore using PLM embeddings for topic discovery. We first identify the challenges with an in-depth analysis of the original PLM embedding space's structure.
\item We propose a new framework \model which jointly learns a lower-dimensional, spherical latent space with cluster structures based on word and document embeddings from PLMs. High-quality topic clusters are derived by simultaneously modeling topic-word and document-topic distributions. \model can be integrated with any PLMs for unsupervised topic discovery.
\item We propose three objectives for training \model to induce distinctive and balanced cluster structures in the latent space which result in diverse and coherent topics.
\item We evaluate \model on two benchmark datasets in different domains. \model significantly outperforms strong topic discovery methods by generating more coherent and diverse topics and providing better document topic representations judged from both automatic and human evaluations.
\end{enumerate}

\section{Challenges of Topic Discovery with Pretrained Language Models}
\label{sec:dilemma}
\begin{figure}[t]
\centering
\subfigure[\textbf{New York Times}.]{
	\label{fig:nyt_vis}
	\includegraphics[width = 0.223\textwidth]{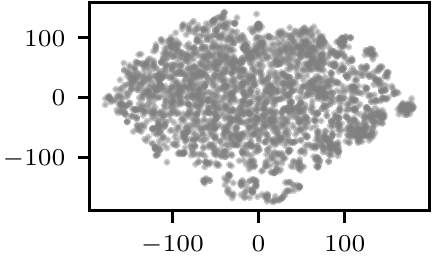}
}
\subfigure[\textbf{Yelp Review}.]{
	\label{fig:yelp_vis}
	\includegraphics[width = 0.223\textwidth]{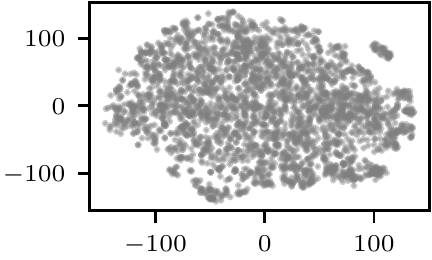}
}
\vspace*{-1em}
\caption{Visualization using t-SNE of $3,000$ randomly sampled contextualized word embeddings of BERT on (a) NYT and (b) Yelp datasets, respectively. The embedding spaces do not have clearly separated clusters.}
\label{fig:challenge}
\vspace*{-1.5em}
\end{figure}

We first identify three major challenges of using PLM embeddings for topic discovery, which motivate our proposed model in Section~\ref{sec:model}.

\noindent
\textbf{Unsuitability of PLM Embedding Space for Clustering.}
One straightforward way of obtaining $K$ topics with PLM embeddings (\eg, from BERT~\cite{devlin2019bert}) is to simply apply clustering algorithms like $K$-means~\cite{lloyd1982least} to group correlated terms that form topics.
To provide empirical evidence that such direct clustering may not work well, we visualize $3,000$ randomly sampled contextualized word embeddings obtained by running BERT on the New York Times and Yelp Review datasets in Figure~\ref{fig:challenge}. The embedding spaces do not exhibit clearly separated clusters, and applying clustering algorithms like $K$-means with a typical $K$ (\eg, $K=100$) to these spaces leads to low-quality and unstable clusters.
We show theoretically that such a phenomenon is due to too many clusters in the embedding space. Below, we study the effect of the Masked Language Modeling (MLM) pretraining objective of BERT on the embedding space.
\begin{theorem}
\label{thm:gmm}
The MLM pretraining objective of BERT assumes that the learned contextualized embeddings are generated from a Gaussian Mixture Model (GMM) with $|V|$ mixture components where $|V|$ is the vocabulary size of BERT.
\end{theorem}
\vspace*{-1em}
\begin{proof}
See Appendix \ref{sec:proof}.
\end{proof}

Theorem~\ref{thm:gmm} applies to many PLMs (\eg, BERT~\cite{devlin2019bert}, RoBERTa~\cite{liu2015topical}, XLNet~\cite{yang2019xlnet}) that use MLM-like pretraining objectives. 
It reveals that the optimal number of cluster $K$ to apply $K$-means like algorithm is $|V|$ ($|V| \approx 30,000$ in the BERT base model). 
In other words, the PLM embedding space is partitioned into extremely fine-grained clusters and lacks topic structures inherently.
If a typical $K$ for topic discovery is used ($K \ll |V|$), the partition will not fit the original data well, resulting in unstable and low-quality clusters. If a very big $K$ is used ($K \approx |V|$), most clusters will contain only one unique term, which is meaningless for topic discovery.

\noindent
\textbf{Curse of Dimensionality.}
PLM embeddings are usually high-dimensional (\eg, number of dimensions $r=768$ in the BERT base model), while distance functions can become meaningless and unreliable in high-dimensional spaces~\cite{beyer1999nearest}, rendering Euclidean distance based clustering algorithms ineffective for high-dimensional cases, known as the ``curse of dimensionality''. 
From another perspective, the high-dimensional PLM embeddings encode linguistic information of multiple aspects for the generic language modeling purpose, but some features are not necessary for or may even interfere with topic discovery. For example, some syntactic features in the PLM embeddings should not be considered when grouping semantically similar concepts (\eg, ``play'', ``plays'' and ``playing'' should not represent different topics).

\noindent
\textbf{Lack of Good Document Representations from PLMs.}
Topic discovery usually requires jointly modeling documents with words to derive latent topics.
Although PLMs are famous for their superior contextualized word representations, obtaining quality document embeddings from PLMs has been a big challenge.
Sentence-BERT~\cite{reimers2019sentence} reports that the inherent BERT sequence embeddings (\ie, obtained from the \texttt{[CLS]} token) are of rather bad quality without fine-tuning, even worse than averaged GloVe context-free embeddings.
To obtain meaningful sentence embeddings, Sentence-BERT fine-tunes pretrained BERT model on natural language inference (NLI) tasks with manually annotated sentences.
However, using Sentence-BERT for topic discovery raises two concerns: (1) When the given corpus has a big domain shift from the Sentence-BERT training set (\eg, the documents are much longer than the sentences in NLI, or are very different semantically from the NLI dataset), the document embeddings need to be re-trained from target corpus document labels, which contradicts the unsupervised nature of topic discovery.
(2) The sentence embeddings are in a different space from word embeddings as they are not jointly trained, and cannot be simultaneously used to model both words and documents. This is why BERTopic~\cite{grootendorst2020bertopic} relies on TF-IDF for topic word selection.

\section{Method}
\label{sec:model}

\begin{figure*}[thb]
\centering
\includegraphics[width = 1.0\textwidth]{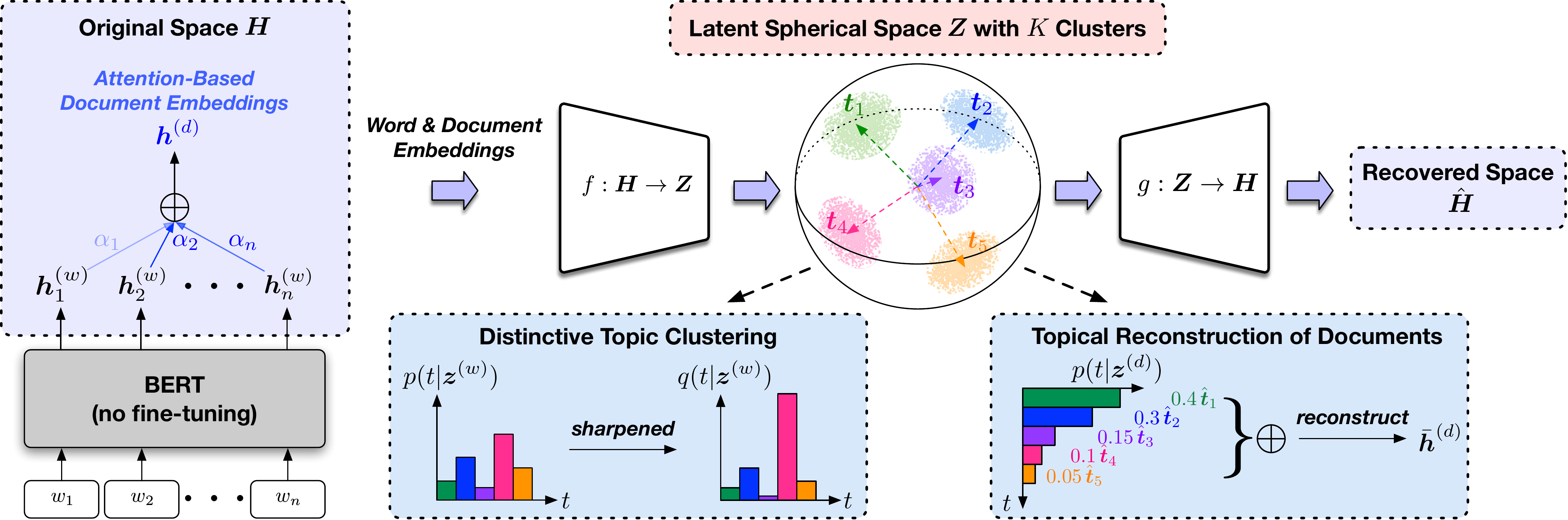}
\vspace*{-2em}
\caption{Overview of \model. We assume that the $K$-topic structure exists in a latent spherical space $\bs{Z}$. We jointly learn the attention weights for document embeddings and the latent space generation model via three objectives: (1) a clustering loss that encourages distinctive topic learning in the latent space, (2) a topical reconstruction loss of documents that promotes meaningful topic representations for summarizing document semantics and (3) an embedding space preserving loss that maintains the semantics of the original embedding space. The PLM is not fine-tuned.}
\label{fig:overview}
\vspace*{-1em}
\end{figure*}

We first introduce the two major components in our \model model: (1) attention-based document embedding learning module and (2) latent space generative module, and then we introduce three training objectives for model learning. 
Figure~\ref{fig:overview} provides an overview of \model.
We assume BERT is used as the PLM, but \model can be seamlessly integrated with any other PLMs.

\subsection{Attention-Based Document Embeddings}
\label{sec:attn}
As the prerequisite of topic discovery is the joint modeling of words and documents, we first propose a simple attention mechanism to learn document embeddings.
Previous studies~\cite{li2020sentence} show that a simple average of word embeddings from PLMs can serve as decent generic sequence representations.
In this work, we assume that not all words in a document are equally topic-indicative, so we learn attention weights of each token to derive document embeddings as a weighted average of contextualized word embeddings which are expected to be better tailored for topic discovery than an unweighted average of word embeddings.
This also allows the learned document embeddings to share the same space with word embeddings which enables joint modeling of words and documents. 

For each text document $\bs{d} = [w_1, w_2, \dots, w_n]$, we obtain the BERT contextualized word representations $[\bs{h}_1^{(w)}, \bs{h}_2^{(w)}, \dots, \bs{h}_n^{(w)}]$ where $\bs{h}_i^{(w)} \in \mathbb{R}^{r}$ ($r = 768$ in the BERT base model).
The attention weights $\bs{\alpha} = [\alpha_1, \alpha_2, \dots, \alpha_n]$ are learned for each token as follows:
$$
\bs{l}_i = \tanh\left(\bs{W}\bs{h}_i^{(w)} + \bs{b}\right),
\quad 
\alpha_i = \frac{\exp(\bs{l}_i^\top \bs{v})}{\sum_{j=1}^n \exp(\bs{l}_j^\top \bs{v})},
$$
where $\bs{W}$ and $\bs{b}$ are learnable parameters of a linear layer with the $\tanh(\cdot)$ activation. Each word embedding $\bs{h}_i^{(w)}$ is transformed to a new representation $\bs{l}_i$ whose dot product with another learnable vector $\bs{v}$ reflects how topic-indicative the token is. Finally, the document embedding $\bs{h}^{(d)}$ is obtained as the combination of all word embeddings in the document weighted by the attention values:
$$
\bs{h}^{(d)} = \sum_{i=1}^n \alpha_i \bs{h}_i^{(w)}.
$$
We note that the contextualized word embeddings from BERT $\{\bs{h}_i^{(w)}\}_{i=1}^n$ are not updated during topic discovery since they already capture word semantics reliably and accurately through pretraining. 
The learnable parameters associated with the attention mechanism $\bs{A} = \{\bs{W}, \bs{b}, \bs{v}\}$ are randomly initialized and trained via the unsupervised objectives to be introduced in Section~\ref{sec:train}.

\subsection{The Latent Space Generative Model}
\label{sec:generative}

\noindent
\textbf{Motivation and Assumptions.}
As we have shown in Section~\ref{sec:dilemma}, the original embedding space $\bs{H}$ of PLMs is unsuitable for direct clustering to generate topic clusters. To address the challenges, we propose to project the original embedding space $\bs{H}$ into a latent space $\bs{Z}$ with $K$ soft clusters of words corresponding to $K$ latent topics. We assume that $\bs{Z}$ is \emph{spherical} (\ie, $\bs{Z} \subset \mathbb{S}^{r'-1}$;  $\mathbb{S}^{r'-1} = \{\bs{z} \in \mathbb{R}^{r'}: \|\bs{z}\| = 1\}$ is the unit $r'-1$ sphere) and \emph{lower-dimensional} (\ie, $r' < r$). Such a latent space has the following preferable properties: 
(1) In the spherical latent space, angular similarity (\ie, without considering vector norms) between vectors is employed to capture word semantic correlations, which works better than Euclidean metrics (\eg, cosine similarity between embeddings is more effective for measuring word similarity~\cite{Meng2019SphericalTE,Meng2020HierarchicalTM}). 
(2) The lower-dimensional space mitigates the ``curse of dimensionality'' of the original high-dimensional space and better suits the clustering task. 
(3) Projecting high-dimensional embeddings to the lower-dimensional space forces the model to discard the information that does not help form topic clusters (\eg, syntactic features).

\noindent
\textbf{Why Not Naive Approach?}
A straightforward way is to first apply a dimensionality reduction technique to the original embedding space $\bs{H}$ to obtain the aforementioned latent space $\bs{Z}$, and subsequently apply clustering algorithms to $\bs{Z}$ for obtaining the latent space clusters representing topics. However, such a naive approach cannot guarantee that the reduced-dimension embeddings will be naturally suited for clustering, given that no clustering-promoting objective is incorporated in the dimensionality reduction step.
Therefore, we propose to \emph{jointly} learn the latent space projection and cluster in the latent space instead of conducting them one after another, so that the latent representation learning is guided by the clustering objective, and the cluster quality benefits from the well-separated structure of the latent space, achieving a mutually-enhanced effect. Such joint learning is realized by training a generative model that connects the latent topic structure with the original space representations.

\noindent
\textbf{Our Generative Model.}
We introduce our latent space generative model as follows. With the number of topics $K$ as the input to the model, we assume that there exists a latent space $\bs{Z} \subset \mathbb{S}^{r'-1}$ with $K$ topics reflecting the latent structure of the original embedding space $\bs{H}$. 
Each topic is associated with a spherical distribution called the von Mises-Fisher (vMF) distribution~\cite{Banerjee2005ClusteringOT,Gopal2014VonMC} that characterizes the topic-word and document-topic distributions in the latent space. 
Specifically, the vMF distribution (can be seen as the spherical counterpart of the Gaussian distribution) of a topic $t$ is parameterized by a mean vector $\bs{t}$ and a concentration parameter $\kappa$. 
The probability density closer to $\bs{t}$ is greater and the spread is controlled by $\kappa$. 
Intuitively, words and documents are more likely to be correlated with a topic $t$ if their latent space representations are closer to the topic vector $\bs{t}$.
Formally, a unit random vector $\bs{z} \in \mathbb{S}^{r'-1}$ has the $r'$-variate vMF distribution $\text{vMF}_{r'}(\bs{t}, \kappa)$ if its probability density function is
\begin{equation*}
\label{eq:vmf}
p(\bs{z};\bs{t}, \kappa) = n_{r'}(\kappa)\exp \left( \kappa \cdot \cos(\bs{z}, \bs{t}) \right),
\end{equation*}
where $\|\bs{t}\| = 1$ is the center direction, $\kappa \ge 0$ is the concentration parameter, $\cos(\bs{z}, \bs{t})$ is the cosine similarity between $\bs{z}$ and $\bs{t}$, and the normalization constant $n_{r'}(\kappa)$ is given by
$$
n_{r'}(\kappa) = \frac{\kappa^{r'/2-1}}{(2\pi)^{r'/2} I_{r'/2-1}(\kappa)},
$$
where $I_{r'/2-1}(\cdot)$ represents the modified Bessel function of the first kind at order $r'/2-1$. 
We assume all topics' vMF distributions share the same concentration parameter $\kappa$ (\ie, the topic terms are equally concentrated around the topic center for all topics) which can be set as a hyperparameter.

Every word embedding $\bs{h}_i^{(w)} \in \bs{H}$ from the original space is 
assumed to be generated through the following process
: (1) A topic $t_k$ is sampled from a uniform distribution over the $K$ topics. (2) A latent embedding $\bs{z}_i^{(w)}$ is generated from the vMF distribution associated with topic $t_k$. (3) A function $g: \bs{Z} \to \bs{H}$ maps the latent embedding $\bs{z}_i^{(w)}$ to the original embedding $\bs{h}_i^{(w)}$ corresponding to word $w_i$. The generative process is summarized as follows:
\begin{equation}
\label{eq:gen_w}
t_k \sim \text{Uniform}(K), \, \bs{z}_i^{(w)} \sim \text{vMF}_{r'}(\bs{t}_k, \kappa), \, \bs{h}_i^{(w)} = g(\bs{z}_i^{(w)}).
\end{equation}

The generative process of document embedding $\bs{h}^{(d)} \in \bs{H}$ is similar since it resides in the same word embedding space:
\begin{equation}
\label{eq:gen_d}
t_k \sim \text{Uniform}(K), \, \bs{z}^{(d)} \sim \text{vMF}_{r'}(\bs{t}_k, \kappa), \, \bs{h}^{(d)} = g(\bs{z}^{(d)}).
\end{equation}

We assume that the mapping function $g$ can be nonlinear to model arbitrary transformations, and we parameterize $g$ as a deep neural network (DNN) since DNNs can approximate any nonlinear function ~\cite{hornik1991approximation}.
Each layer $l$ in the DNN is a linear layer with ReLU activation function, taking $\bs{x}_l$ as input and outputting $\bs{y}_l$:
$$
\bs{y}_l = \text{ReLU} (\bs{W}_l \bs{x}_l + \bs{b}_l),
$$
where $\bs{W}_l$ and $\bs{b}_l$ are the learnable parameters in the layer.
We also jointly learn the mapping $f: \bs{H} \to \bs{Z}$ from the original space to the latent space (\ie, the inverse function of $g$, also parameterized by a DNN) to map unseen word/document embeddings to the latent space. Such joint learning of two nonlinear functions follows an autoencoder~\cite{hinton1994autoencoders} setup where an encoding network maps data points from the original space to the latent space, and a decoding network converts latent space data back to an approximate reconstruction of the original data. 
\subsection{Model Training}
\label{sec:train}
\begin{figure*}[t]
\centering
\subfigure[Start of EM Algorithm.]{
	\includegraphics[width = 0.32\textwidth]{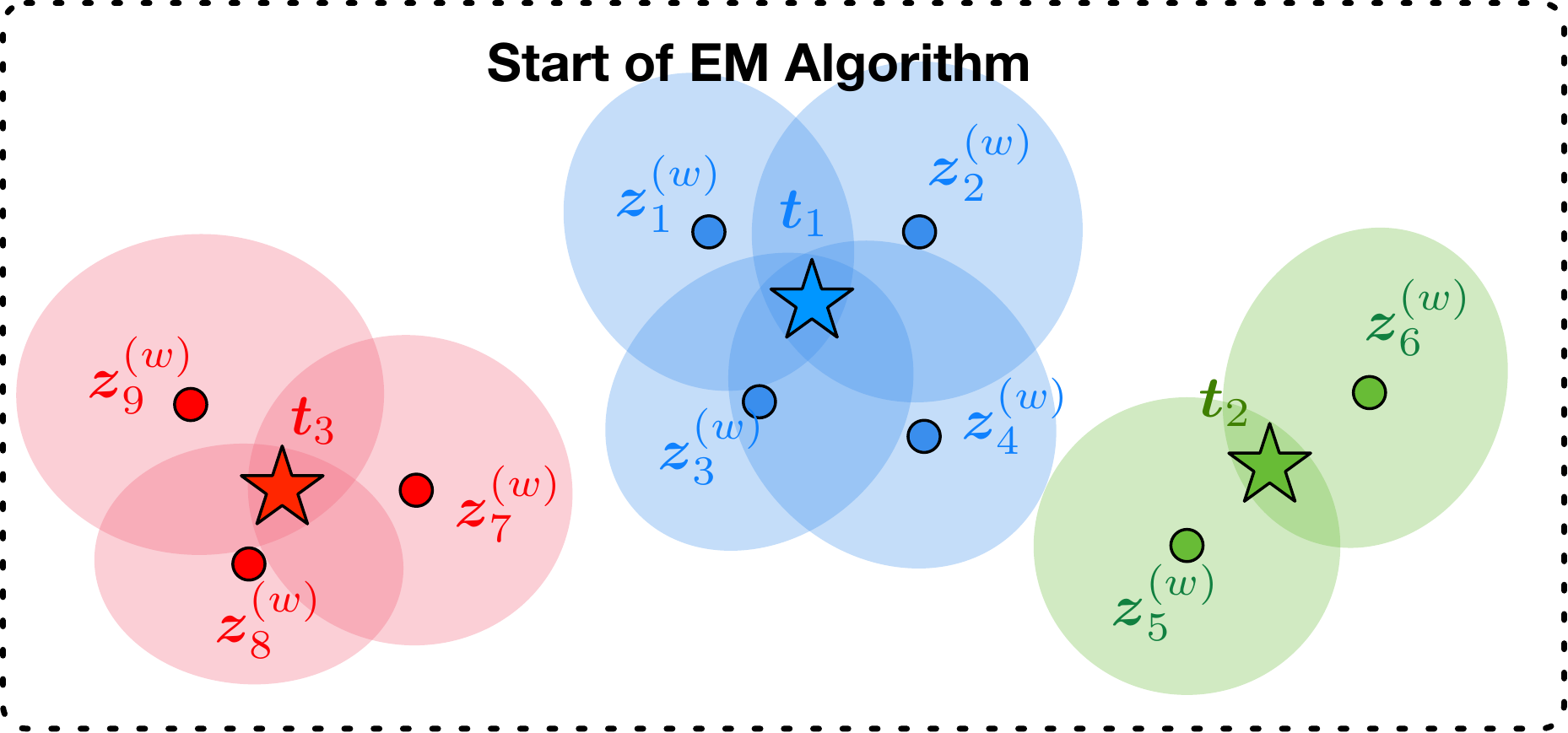}
}
\subfigure[E-Step.]{
	\includegraphics[width = 0.32\textwidth]{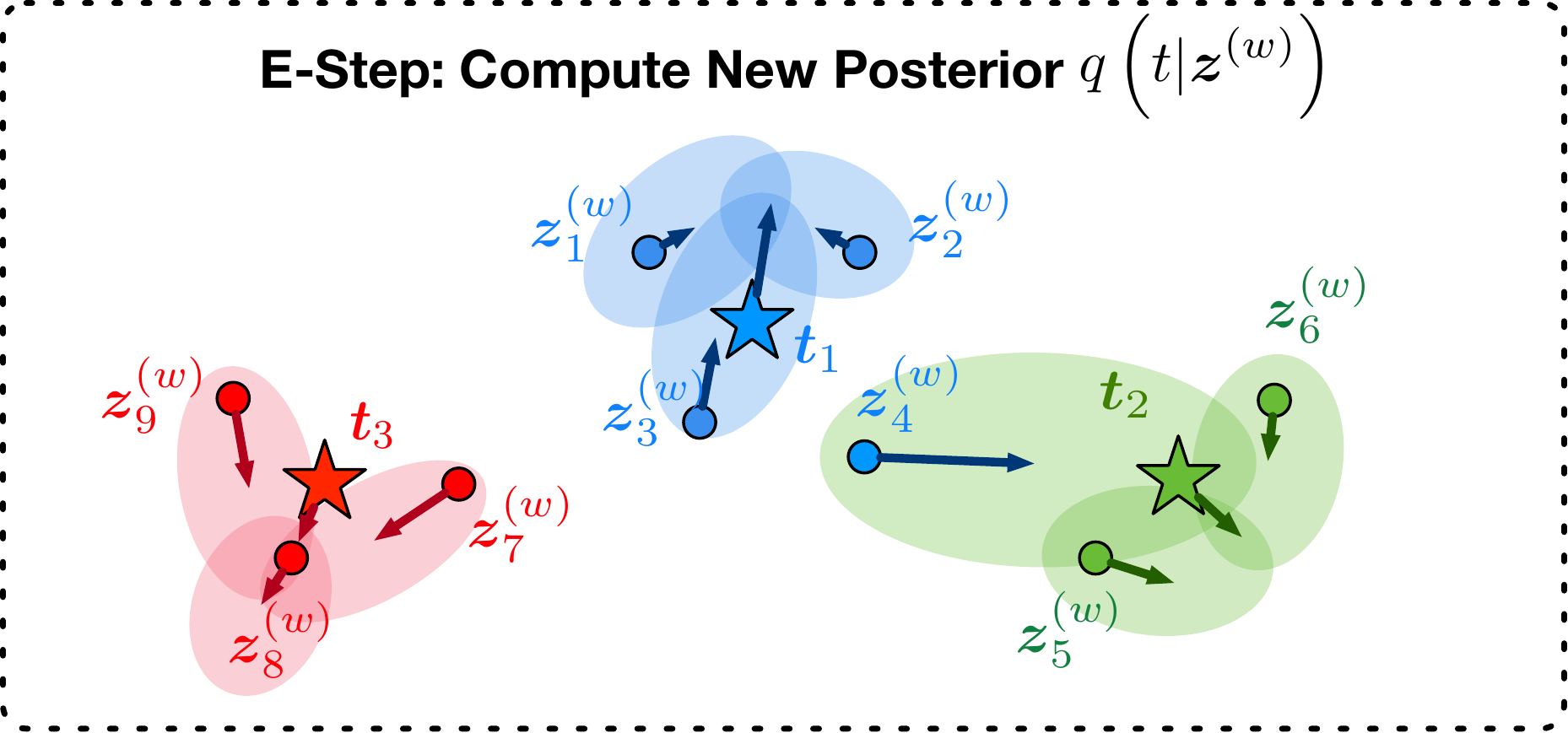}
}
\subfigure[M-Step.]{
	\includegraphics[width = 0.32\textwidth]{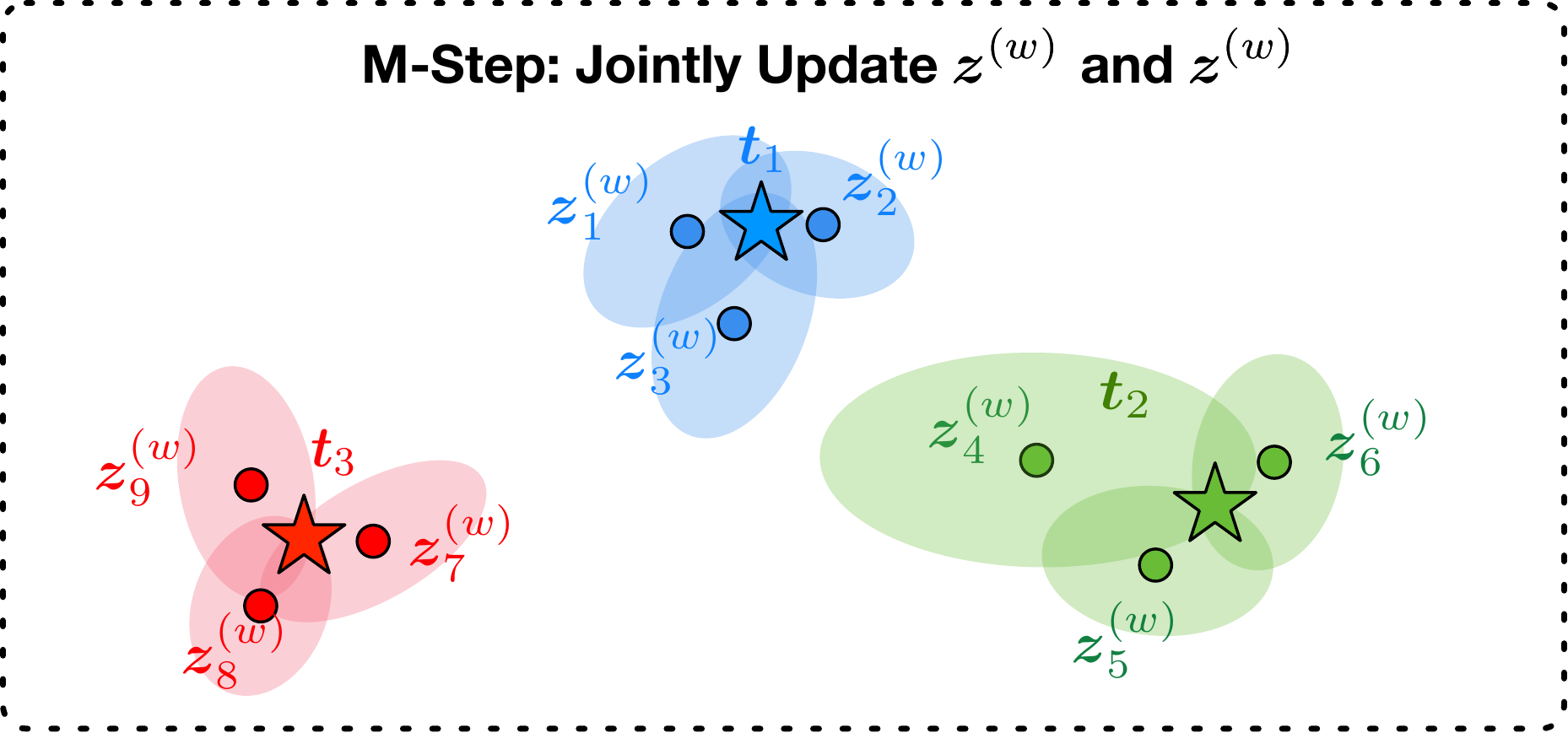}
}
\vspace*{-1.5em}
\caption{One iteration of EM algorithm. During the E-Step, we compute new posterior topic-word distribution $q(t|\bs{z}^{(w)})$ that sharpens the original posterior $p(t|\bs{z}^{(w)})$ (resulting in lower entropy of $p(t|\bs{z}^{(w)})$ denoted by the smaller colored area around $\bs{z}^{(w)}$) and meanwhile encourage balanced cluster distribution (resulting in some cluster assignment change). During the M-Step, we update topic embeddings $\bs t$ and word embeddings ${\bs z^{(w)}} = f({\bs h^{(w)}})$ according to the new posteriors.}
\label{fig:em}
\vspace*{-1em}
\end{figure*}

To jointly train the attention module for document embeddings in Section~\ref{sec:attn} and the latent generative model in Section~\ref{sec:generative}, we introduce three objectives: (1) a clustering loss that enforces separable cluster structures in the latent space for distinctive topic learning, (2) a topical reconstruction loss of documents to ensure the discovered topics are meaningful summaries of document semantics, and (3) an embedding space preserving loss to maintain the semantic information in the original space.

\noindent
\textbf{Distinctive Topic Clustering.}
The first clustering objective induces a latent space with $K$ well-separated clusters by gradually sharpening the posterior topic-word distributions via an expectation–maximization (EM) algorithm. In the E-step, we estimate a new (soft) cluster assignment of each word based on the current parameters; in the M-step, we update the model parameters given the cluster assignments. The process is illustrated in Figure~\ref{fig:em}.

\noindent
\textit{E-Step.} To estimate the cluster assignment of each word, we compute the posterior topic distribution obtained via the Bayes rule:
$$
p\left(t_k \big| \bs{z}_i^{(w)}\right) = \frac{p\left(\bs{z}_i^{(w)} \big| t_k\right) p(t_k)}{\sum_{k'=1}^{K} p\left(\bs{z}_i^{(w)} \big| t_{k'}\right) p(t_{k'})},
$$
where $p\left(\bs{z}_i^{(w)} | t_k\right) = \text{vMF}_{r'}(\bs{t}_k, \kappa) = n_{r'}(\kappa)\exp \left( \kappa \cdot \cos(\bs{z}_i^{(w)}, \bs{t}_k) \right)$ and $p(t_k) = 1/K$ according to Eq.~\eqref{eq:gen_w}. The posterior is simplified as
\begin{equation*}
p\left(t_k \big| \bs{z}_i^{(w)}\right) = \frac{\exp \left(\kappa \cdot \cos\left(\bs{z}_i^{(w)}, \bs{t}_k\right) \right)}{\sum_{k'=1}^{K} \exp \left(\kappa \cdot \cos\left(\bs{z}_i^{(w)}, \bs{t}_{k'}\right) \right)}.
\end{equation*}
Then we compute a new estimate of the cluster assignments $q(t_k | \bs{z}_i^{(w)})$ to be used for updating the model in the M-Step following \cite{xie2016unsupervised}: 
\begin{equation}
\label{eq:target}
q\left(t_k \big| \bs{z}_i^{(w)}\right) = \frac{p\left(t_k \big| \bs{z}_i^{(w)}\right)^2/s_k}{\sum_{k'=1}^{K} p\left(t_{k'} \big| \bs{z}_i^{(w)}\right)^2/s_{k'}},
\,
s_{k}=\sum_{i=1}^{N} p\left(t_k \big| \bs{z}_i^{(w)}\right),
\end{equation}
where $N$ is the total number of tokens in the corpus.
Using Eq.~\eqref{eq:target} to obtain the target cluster assignment has the following two favorable effects: (1) \emph{Distinctive topic learning}. Squaring-then-normalizing the posterior distribution $p(t_k | \bs{z}_i^{(w)})$ has a sharpening effect that skews the distribution towards its most confident cluster assignment, and the so learned latent space will have gradually well-separated clusters for distinctive topic interpretation. 
This is similar in spirit to the Dirichlet prior used in LDA that promotes sparse topic distributions.
(2) \emph{Topic prior regularization}. The soft cluster frequency $s_{k}$ should encode the uniform topic prior assumed in Eq.~\eqref{eq:gen_w}, and dividing the sharpened $p(t_k | \bs{z}_i^{(w)})^2$ by $s_{k}$ encourages balanced clusters.

\noindent
\textit{M-Step.} We update the model parameters to maximize the expected log-probability of the current cluster assignment under the new cluster assignment estimate $\mathbb{E}_q[\log p]$, which is equivalent to minimizing the following cross entropy loss:
\begin{equation}
\label{eq:cluster}
\mathcal{L}_{\text{clus}} = - \sum_{i=1}^{N} \sum_{k=1}^{K} q\left(t_k \big| \bs{z}_i^{(w)}\right) \log p\left(t_k \big| \bs{z}_i^{(w)}\right),
\end{equation}
where $p$ is updated to approximate $q$ which is a fixed target. Using Eq.~\eqref{eq:cluster} to update the model parameters has a notable difference from standard clustering algorithms: Since $p(t_k | \bs{z}_i^{(w)})$ is jointly determined by the topic center vector $\bs{t}_k$ and latent representation $\bs{z}_i^{(w)}$, both of them will be updated to fit the new estimate $q(t_k | \bs{z}_i^{(w)})$ which encourages distinctive cluster distribution. Therefore, the mapping function $f$ will be adjusted accordingly to induce a latent space with a $K$-cluster structure and the topic center vectors will become $K$ anchoring points surrounded by topic-representative words. In contrast, standard clustering algorithms 
only update the cluster parameters without changing the data representations.

\noindent
\textbf{Topical Reconstruction of Documents.}
The second objective aims to reconstruct document semantics with topic representations so that the learned latent topics are meaningful summaries of the documents.
Specifically, the reconstructed document embedding $\hat{\bs{h}}^{(d)}$ is obtained by combining all projected topic vectors $\hat{\bs{t}}_{k}$ weighted by the document-topic distribution $p(t_k | \bs{z}^{(d)})$:
$$
\hat{\bs{h}}^{(d)} = \sum_{k=1}^{K} p\left(t_k \big| \bs{z}^{(d)}\right) \hat{\bs{t}}_{k}, 
\quad
\hat{\bs{t}}_{k} = g(\bs{t}_{k}),
$$
where $p(t_k | \bs{z}^{(d)})$ is obtained according to Eq.~\eqref{eq:gen_d}:
$$
p\left(t_k \big| \bs{z}^{(d)}\right) = \frac{\exp \left(\kappa \cdot \cos\left(\bs{z}^{(d)}, \bs{t}_k\right) \right)}{\sum_{k'=1}^{K} \exp \left(\kappa \cdot \cos\left(\bs{z}^{(d)}, \bs{t}_{k'}\right) \right)}.
$$

We require the reconstructed document embedding to be a good approximation of the original content by minimizing the following reconstruction loss:
\begin{equation}
\label{eq:doc_recons}
\mathcal{L}_{\text{rec}} = \sum_{d \in \mathcal{D}} \big\|\hat{\bs{h}}^{(d)} - \bar{\bs{h}}^{(d)}\big\|^2,
\end{equation}
where $\bar{\bs{h}}^{(d)}$ is the average of word embeddings in the document serving as the generic document embedding.

\noindent
\textbf{Preservation of Original PLM Embeddings.}
We need to ensure the latent space preserves the important semantic information of the original embedding space, and the third objective 
encourages 
the output of the autoencoder to faithfully recover the structure of the original embedding space by minimizing the the following loss:
\begin{equation}
\label{eq:recons}
\mathcal{L}_{\text{pre}} = \sum_{i=1}^N \big\|\bs{h}_i^{(w)} - g \left( f\left(\bs{h}_i^{(w)}\right)\right) \big\|^2.
\end{equation}



\noindent
\textbf{Overall Algorithm.}
We summarize the training of \model in Algorithm~\ref{alg:train}. 
We first pretrain the mapping functions $f$ and $g$ only using the preservation loss in Eq.~\eqref{eq:recons} as it provides a stable initialization of the latent space~\cite{xie2016unsupervised}. During training, we apply the EM algorithm to iteratively update all model parameters with the summed objectives (the clustering loss is weighed by $\lambda$). 
\SetKwInput{KwParameter}{Parameter}
\SetKwInput{KwHyperParameter}{Hyperparameter}
\begin{algorithm}[t]
\caption{\model Training.}
\label{alg:train}
\KwIn{
$\mathcal{D}$: Text corpus; $M$: PLM; $K$: Number of topics.
}
\KwParameter{
$\bs{A}$: Attention mechanism parameters; $f,g$: Encoding/decoding functions; $\bs{T}$: Topic embeddings.
}
\KwHyperParameter{
$E$: Training epochs; $\lambda$: Clustering loss weight.
}
\KwOut{Topic-word distributions $p\left(z_i^{(w)}\big|t_k\right)$; document-topic distributions $p\left(t_k\big|z^{(d)}\right)$.}
$f, g \gets \arg\min_{f,g} \mathcal{L}_{\text{pre}};$ // Pretrain $f,g$ via Eq.~\eqref{eq:recons}\;
$\bs{T} = \bs{t}_k\big\rvert _{k=1}^{K} \gets$ Initialize with $K$-means on $\mathbb{S}^{r'-1}$\;
\For{$j \in [1, 2, \dots, E]$}  {
// E-Step: Update cluster assignment estimation\;
$q\left(t_k \big| \bs{z}_i^{(w)}\right) \gets $ Eq.~\eqref{eq:target}\;
// M-Step: Update model parameters\;
$\bs{A}, f, g, \bs{T} \gets \arg\min_{\bs{A},f,g,\bs{T}}   \left(\lambda\mathcal{L}_{\text{clus}} + \mathcal{L}_{\text{rec}} + \mathcal{L}_{\text{pre}}\right) $;
}
\Return $p\left(z_i^{(w)}\big|t_k\right), p\left(t_k\big|z^{(d)}\right)$\;
\end{algorithm}

\noindent
\textbf{Complexity.} 
In the E-Step of the algorithm, $q(t_k | \bs{z}_i^{(w)})$ is computed for every latent representation over each topic, resulting in an $\mathcal{O}(NKr')$ complexity per iteration. The M-Step updates DNN parameters whose complexity is related to the number of parameters in the model and the optimization method.



\section{Experiments}

\subsection{Experiment Setup}
\noindent
\textbf{Settings.}
We use two benchmark datasets in different domains with long/short texts for evaluation: (1) The New York Times annotated corpus (\textbf{NYT})~\cite{Sandhaus2008}; and (2) The Yelp Review Challenge dataset
(\textbf{Yelp}). The dataset statistics can be found in Table~\ref{tab:dataset_stats}.
The implementation details and parameters of \model are shown in Appendix \ref{sec:hyperparameter}.
For both datasets, we set the number of topics $K = 100$ for all compared methods.

\noindent
\textbf{Compared Methods.} We compare \model with the following strong baselines:
\begin{itemize}[wide, labelwidth=!, labelindent=0pt] 
\item LDA~\cite{Blei2003LatentDA}: LDA is the standard topic model that learns topic-word and document-topic distributions by modeling the generative process of the corpus.
\item CorEx~\cite{gallagher2017anchored}: CorEx does not rely on generative assumptions and learns maximally informative topics measured by total correlation.
\item ETM~\cite{dieng2020topic}: ETM models word topic correlations via distributed representations to improve the expressiveness of topic models.
\item BERTopic~\cite{grootendorst2020bertopic}: BERTopic first clusters document embeddings from BERT and then uses TF-IDF to extract topic representative words, which does not leverage word embeddings from PLMs.
\end{itemize}

\subsection{Topic Discovery Evaluation}
\noindent
\textbf{Evaluation Metrics.}
We evaluate the quality of the topics from two aspects: \emph{topic coherence} and \emph{topic diversity}. Good topic results should be both coherent for humans to interpret and diverse to cover more information about the corpus. We evaluate the effectiveness of document-level topic modeling by document clustering. 

\noindent
For topic coherence, we use three metrics including both human and automatic evaluations:
\begin{itemize}[wide, labelwidth=!, labelindent=0pt] 
\item UMass~\cite{mimno2011optimizing}: UMass computes the log-conditional probability of every top word in each topic given every other top word that has a higher order in the ranking of that topic. 
The probability is computed based on document-level word co-occurrence.
\item UCI~\cite{newman2010automatic}: UCI computes the average pointwise mutual information of all pairs of top words in each topic. The word co-occurrence counts are derived using a sliding window of size $10$.
\item Intrusion: Given the top terms of a topic, we inject an intrusion term that is randomly chosen from another topic. Then a human evaluator is asked to identify the intruded term. The more coherent the top terms are, the more likely an evaluator can correctly identify the fake term, and thus we compute the ratio of correctly identified intrusion instances as the topic coherence score given by the intrusion test. The topics from all compared methods are randomly shuffled during evaluation to avoid the bias of human evaluators.
\end{itemize}
\noindent
For topic diversity, we report the percentage of unique words in the top words of all topics following the definition in \cite{dieng2020topic}.

\noindent
\textbf{Qualitative Evaluation.}
We randomly select several ground truth topics from both datasets, and manually match the most relevant  topic generated by all methods. Table~\ref{tab:quality} shows the top-$5$ words per topic. All methods are able to generate relevant topics to the ground truth ones. LDA and CorEx results contain noises that are semantically irrelevant to the topic; ETM improves LDA by incorporating word embeddings, but still generates slightly off-topic terms; BERTopic also has noisy terms in the results, as it uses TF-IDF metrics without exploiting word representations from BERT for obtaining top words. \model consistently outputs coherent and meaningful topics.
\setlength{\tabcolsep}{3pt}
\begin{table*}[ht]
\small
\centering
\caption{Qualitative evaluation of topic discovery. We select several ground truth topics and manually find the most relevant topic generated by all methods. Words not strictly belonging to the corresponding topic are italicized and underlined.}
\vspace*{-1em}
\label{tab:quality}
\resizebox{\textwidth}{!}{
\begin{tabular}{c|ccccc|ccccc}
\toprule
\multirow{3}{*}{Methods} &
\multicolumn{5}{c|}{\textbf{NYT}} &
\multicolumn{4}{c}{\textbf{Yelp}} \\
& Topic 1 & Topic 2 & Topic 3 & Topic 4 & Topic 5 & Topic 1 & Topic 2 & Topic 3 & Topic 4 & Topic 5\\
 & (sports) & (politics) & (research) & (france) & (japan) & (positive) & (negative) & (vegetables) & (fruits) & (seafood) \\
\midrule
\multirow{5}{*}{\makecell{LDA}}  & olympic & \wrong{mr} & \wrong{said} & french & japanese & amazing & loud & spinach & mango & fish\\
 & \wrong{year} & bush & report & \wrong{union} & tokyo & \wrong{really} & awful & carrots & strawberry & \wrong{roll}\\
 & \wrong{said} & president & evidence & \wrong{germany} & \wrong{year} & \wrong{place} & \wrong{sunday} & greens& \wrong{vanilla} & salmon\\
 & games & white & findings & \wrong{workers} & matsui & phenomenal  &  \wrong{like} & salad & banana & \wrong{fresh}\\
& team & house & defense & paris & \wrong{said} & pleasant  & slow & \wrong{dressing} & \wrong{peanut} & \wrong{good}\\

\midrule
\multirow{5}{*}{\makecell{CorEx}}  & baseball & house & possibility & french & japanese & great & \wrong{even} & garlic & strawberry & shrimp\\
 & championship & white & challenge & \wrong{italy} & tokyo & friendly & bad & tomato & \wrong{caramel} & \wrong{beef}\\
 & playing & support & reasons & paris & \wrong{index} & \wrong{atmosphere} & mean & onions & \wrong{sugar} & crab\\
 & \wrong{fans} & \wrong{groups} & \wrong{give} & francs & osaka & love & cold & \wrong{toppings} & fruit & \wrong{dishes}\\
& league & \wrong{member} & planned & jacques & \wrong{electronics} & favorite & \wrong{literally} & \wrong{slices} & mango & \wrong{salt}\\

\midrule
\multirow{5}{*}{\makecell{ETM}}  & olympic & government & approach & french & japanese & nice  &  disappointed & avocado & strawberry & fish\\
& league & national  & problems & \wrong{students} & \wrong{agreement} & worth  & cold & \wrong{greek} & mango & shrimp\\
 & \wrong{national} & \wrong{plan} & experts & paris & tokyo &  \wrong{lunch} & \wrong{review} & salads & \wrong{sweet} & lobster\\
& basketball & public & \wrong{move} & \wrong{german} & \wrong{market} & recommend  & \wrong{experience} & spinach & \wrong{soft} & crab\\
 & athletes & support & \wrong{give} & \wrong{american} & \wrong{european} & friendly  &  bad & tomatoes & \wrong{flavors} & \wrong{chips}\\
\midrule

\multirow{5}{*}{\makecell{BERTopic}}  & swimming & bush & researchers & french & japanese & awesome  & horrible & tomatoes & strawberry & lobster\\
 & freestyle & democrats & scientists & paris & tokyo &  \wrong{atmosphere} & \wrong{quality} & avocado & mango & crab\\
 & \wrong{popov} & white & cases & lyon & ufj &  friendly & disgusting & \wrong{soups} & \wrong{cup} & shrimp\\
& gold & bushs & \wrong{genetic} & \wrong{minister} & \wrong{company} & \wrong{night}  & disappointing & kale & lemon & oysters\\
 & olympic & house & study & \wrong{billion} & yen &  good & \wrong{place} & cauliflower & banana & \wrong{amazing}\\
\midrule

\multirow{5}{*}{\makecell{\model}} & athletes & government & hypothesis & french & japanese & good & tough & potatoes & strawberry & fish \\
 & medalist & ministry & methodology & seine & tokyo & best & bad & onions & lemon & octopus\\
 & olympics & bureaucracy & possibility & toulouse & osaka & friendly & painful & tomatoes & apples & shrimp\\
 & tournaments & politicians & criteria & marseille & hokkaido & cozy & frustrating & cabbage & grape & lobster\\
& quarterfinal & electoral & assumptions & paris & yokohama & casual & brutal & mushrooms & peach & crab\\

\bottomrule

\end{tabular}
}
\vspace*{-1em}
\end{table*}
\setlength{\tabcolsep}{6pt}

\noindent
\textbf{Quantitative Evaluation.}
We report the performance of all methods under the four metrics in Table~\ref{tab:quantity}. Overall, the quantitative evaluation coincides with the previous qualitative results. \model generates not only the most coherent but also diverse topics, under both automatic and human evaluations.

\setlength{\tabcolsep}{3pt}
\begin{table}[thb]
\centering
\caption{Quantitative evaluation of topic discovery. We evaluate all methods with three topic coherence metrics UCI, UMAss and Intrusion (Int.) and a topic diversity (Div.) metric. Higher score means better for all metrics. We do not report Div. for CorEx because it requires topics to have non-overlapping words by design.}
\vspace*{-1em}
\label{tab:quantity}
\scalebox{1.0}{
\begin{tabular}{c|cccc|cccc}
\toprule
\multirow{2}{*}{Methods} &
\multicolumn{4}{c|}{\textbf{NYT}} &
\multicolumn{4}{c}{\textbf{Yelp}} \\
& UMass & UCI & Int. & Div. & UMass & UCI & Int. & Div. \\
\midrule
LDA & -3.75 & -1.76 & 0.53 & 0.78 & -4.71 & -2.47 & 0.47 & 0.65 \\
CorEx & -3.83 & -0.96 & 0.77 & - & -4.75 & -1.91 & 0.43 & - \\
ETM & -2.98 & -0.98 & 0.67 & 0.30 & -3.04 & -0.33 & 0.47 & 0.16 \\
BERTopic & -3.78 & -0.51 & 0.70 & 0.61 & -6.37 & -2.05 & 0.73 & 0.36 \\
\model & \textbf{-2.67} & \textbf{-0.45} & \textbf{0.93} & \textbf{0.99} & \textbf{-1.35} & \textbf{-0.27} & \textbf{0.87} & \textbf{0.96} \\
\bottomrule
\end{tabular}
}
\vspace*{-1em}
\end{table}
\setlength{\tabcolsep}{6pt}




\subsection{Document Clustering Evaluation}

\begin{table}[thb]
\small
\centering
\caption{Document clustering NMI scores on \textbf{NYT} (Topic/Location label set).}
\vspace*{-1em}
\label{tab:clustering}
\resizebox{\columnwidth}{!}{
\begin{tabular}{ccccc}
\toprule
LDA &CorEx & ETM & BERTopic & \model \\
\midrule
0.39/0.20 & 0.29/0.20 & 0.41/0.21 & 0.26/0.22 & \textbf{0.46}/\textbf{0.28} \\
\bottomrule
\end{tabular}
}
\vspace*{-1em}
\end{table}

\begin{figure*}[t]
\centering
\subfigure[Epoch 0.]{
	\includegraphics[width = 0.225\textwidth]{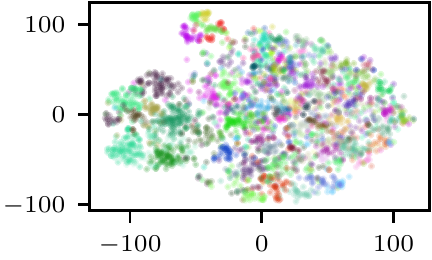}
}
\subfigure[Epoch 2.]{
	\includegraphics[width = 0.225\textwidth]{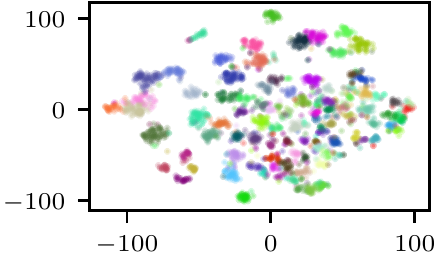}
}
\subfigure[Epoch 4.]{
	\includegraphics[width = 0.225\textwidth]{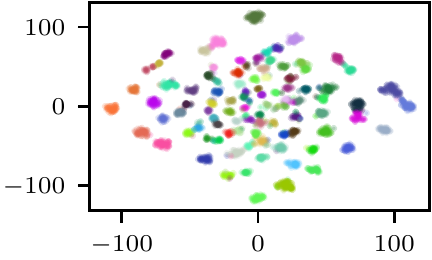}
}
\subfigure[Epoch 8.]{
	\includegraphics[width = 0.225\textwidth]{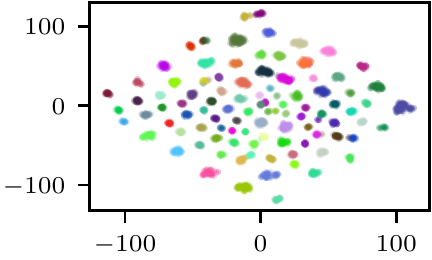}
}
\vspace*{-1em}
\caption{Visualization using t-SNE of $3,000$ randomly sampled latent word embeddings during training. Embeddings assigned to the same cluster are in the same color. The latent space gradually exhibits distinctive and balanced cluster structure.}
\label{fig:vis}
\vspace*{-1em}
\end{figure*}

\noindent
\textbf{Evaluation Metrics.}
We use the learned latent document embedding $\bs{z}^{(d)}$ as the feature to $K$-Means for obtaining document clusters, then we report the Normalized Mutual Information (NMI) score between the clustering results and the ground truth document labels. 

We use the topic label set (\eg, politics, sports) and location label set (\eg, United States, China) on the \textbf{NYT} dataset. The detailed label statistics can be found in \cite{meng2020discriminative}.
On the two label sets, the document-topic distribution learned by \model consistently yields the best clustering results among all methods as shown in Table~\ref{tab:clustering}.

\subsection{Study of \model Training}
\begin{figure}[tbh]
\centering
\subfigure[Topic Quality.]{
    \label{fig:quality}
	\includegraphics[width = 0.225\textwidth]{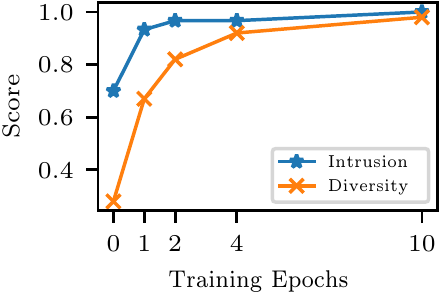}
}
\subfigure[Document Clustering.]{
    \label{fig:clustering}
	\includegraphics[width = 0.225\textwidth]{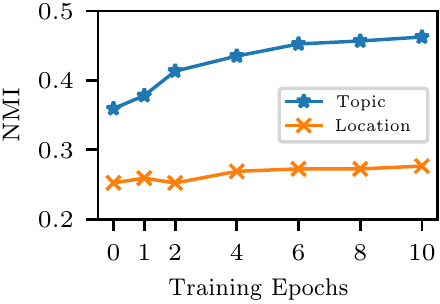}
}
\vspace*{-1em}
\caption{Study of \model training on \textbf{NYT}. We show (a) topic coherence measured by intrusion test and topic diversity and (b) document clustering NMI scores over training.}
\label{fig:training}
\vspace*{-2em}
\end{figure}

\noindent
\textbf{Joint Learning Latent Space and Clustering Improves Topic Quality.} 
Figure~\ref{fig:training} shows the improvement in topic quality (measured by both intrusion test score and topic diversity) and document clustering performance during \model training. At epoch $0$, the result is equivalent to first applying dimensionality reduction (\ie, pretraining autoencoder with $\mathcal{L}_{\text{pre}}$) and then clustering with $K$-means, the ``naive approach'' mentioned in the second paragraph of Section~\ref{sec:generative}. 
Its inferior performance confirms that conducting the two steps separately does not generate satisfactory topics. 
Topic quality and document clustering performance improve when the model is trained longer, showing that joint latent space learning and clustering indeed helps generate coherent and distinctive topics.

\noindent
\textbf{Visualization.}
To intuitively understand how \model jointly learns the latent space structure and performs clustering, we visualize the learned latent embeddings at different training epochs in Figure~\ref{fig:vis}. Before the training starts (epoch $0$), the latent embedding space does not have clear cluster structures, just like the original space. During training, the latent embeddings are becoming well-separated and the cluster structure is gradually more distinctive and balanced, resulting in coherent and diverse topics.

\section{Related Work}

\subsection{Topic Models}
Topic models aim to discover underlying topics and semantic structures from text corpora. Despite extensive studies of topic models following LDA, most approaches suffer from one or more of the following limitations: (1) \textit{The ``bag-of-words'' assumption} that presumes words in the document are generated independently from each other. 
(2) \textit{The reliance on local corpus statistics}, which could be improved by leveraging general knowledge such as pretrained language models~\cite{devlin2019bert}.
(3) \textit{The intractable posterior} that requires approximation techniques during model inference.

Topic modeling approaches can be divided into three major categories: (1) \textit{LDA-based approaches} use pLSA~\cite{hofmann2004latent} or LDA~\cite{Blei2003LatentDA} as the backbone. The idea is to characterize documents as mixtures of latent topics and represent each topic as a distribution over words. Popular models in this category include Hierarchical LDA~\cite{griffiths2004hierarchical}, Dynamic Topic Models~\cite{blei2006dynamic}, Correlated Topic Models~\cite{blei2006correlated}, Pachinko Allocation~\cite{li2006pachinko}, Supervised Topic Models~\cite{mcauliffe2008supervised} and Labeled LDA~\cite{ramage2009labeled}.
Most of these models suffer from all three limitations mentioned above. (2) \textit{Topic models with word embeddings} have been broadly studied after word2vec~\cite{Mikolov2013DistributedRO} came out. The common strategy is to convert the discrete text into continuous representations of embeddings, and then adapt LDA to generate real-valued data. Such kind of models include Gaussian LDA~\cite{das2015gaussian}, Spherical Hierarchical Dirichlet Process~\cite{batmanghelich2016nonparametric} and WELDA~\cite{Bunk2018WELDAET}. There are some other strategies combining topic modeling and word embedding. For example, LFTM~\cite{nguyen2015improving} models a mixture of the multinomial distribution and a link function between word and topic embeddings. TWE~\cite{liu2015topical} uses pretrained topic structures to learn topic embeddings and improve word embeddings.
Although these models consider word embeddings to make up for the ``bag-of-words'' assumption, they are not equipped with general knowledge from pretrained language models. (3) \textit{Neural topic models} are inspired by deep generative models such as VAE~\cite{kingma2013auto}. 
NVDM~\cite{miao2016neural} encodes documents with variational posteriors in the latent topic space. Instead, ProdLDA~\cite{srivastava2017autoencoding} proposes a Laplace approximation of Dirichlet distributions to enable reparameterization. 
Although these neural topic models improve the posterior approximation with neural networks, they still do not utilize general knowledge such as pretrained language models.




\subsection{Pretrained Language Models}
Bengio et al.~\cite{bengio2003neural} propose the Neural Network Language Model which pioneers the study of modern word embedding. Mikolov et al.~\cite{Mikolov2013DistributedRO} introduce two architectures, CBOW and Skip-Gram, to capture local context semantics of each word. 

Although word embeddings have been shown effective in NLP tasks, they are context-independent. Meanwhile, most NLP tasks are beyond word-level, thus it is beneficial to derive word semantics based on specific contexts. Therefore, contextualized PLMs are widely studied recently. 
For example, 
BERT~\cite{devlin2019bert} and RoBERTa~\cite{liu2019roberta} adopt masked token prediction as the pretraining task to leverage bidirectional contexts. 
XLNet~\cite{yang2019xlnet} proposes a new pretraining objective on a random permutation of input sequences. 
ELECTRA~\cite{clark2020electra}, COCO-LM~\cite{meng2021coco} and AMOS~\cite{meng2022pretraining} use a generator to replace some tokens of a sequence and predict whether a token is replaced given its surrounding context. 
For more related studies, one can refer to a recent survey~\cite{qiu2020pre}. 
There have been a few recent studies that attempt to incorporate PLM representations into the topic modeling framework for different purposes~\cite{Bianchi2021PretrainingIA,Chaudhary2020TopicBERTFE,Gupta2021MultisourceNT,Hoyle2020ImprovingNT,Thompson2020TopicMW}.
By contrast, our approach features a latent space clustering framework that leverages the inherent representations of PLMs for topic discovery without following the topic modeling setup.

\section{Conclusion}
We explore a new alternative to topic models via latent space clustering of PLM representations. 
We first analyze the challenges of using PLM embeddings to generate topic structures, and then propose a joint latent space learning and clustering approach \model to address the identified challenges.
\model generates coherent and distinctive topics and outperforms strong topic modeling baselines in both topic quality and topical document representations.
We also conduct studies to provide insights on how the joint learning setup in \model gradually improves the generated topic quality. 

\model is conceptually simple which facilitates future extensions such as integrating with new PLMs and advanced clustering techniques. \model may also be extended to perform hierarchical topic discovery, perhaps via top-down clustering in the latent space. Other related tasks like taxonomy construction~\cite{Huang2020CoRel} and weakly-supervised text classification~\cite{Huang2020WeaklySupervisedAS,Meng2018WeaklySupervisedNT,Meng2019WeaklySupervisedHT,meng2020weakly,zhang2022motifclass} may benefit from the coherent and distinctive topics generated by \model.
\end{spacing}

\begin{acks}
Research was supported in part by US DARPA KAIROS Program No.\ FA8750-19-2-1004, SocialSim Program No.\ W911NF-17-C-0099, and INCAS Program No.\ HR001121C0165, National Science Foundation IIS-19-56151, IIS-17-41317, and IIS 17-04532, and the Molecule Maker Lab Institute: An AI Research Institutes program supported by NSF under Award No.\ 2019897. Any opinions, findings, and conclusions or recommendations expressed herein are those of the authors and do not necessarily represent the views, either expressed or implied, of DARPA or the U.S. Government.
Yu Meng is supported by the Google PhD Fellowship.
We thank anonymous reviewers for valuable and insightful feedback.
\end{acks}

\balance
\bibliographystyle{ACM-Reference-Format}
\bibliography{ref}

\clearpage
\appendix
\section{Ethical Considerations}
PLMs have been shown to contain potential biases~\cite{Prabhumoye2018StyleTT} which may be carried to the downstream applications.
Our work focuses on using representations from PLMs for discovery of topics in a target corpus, and the results will be related to both the PLMs and the corpus statistics.
We suggest applying our method together with bias reduction and correction techniques for PLMs~\cite{Gehman2020RealToxicityPromptsEN,Ma2020PowerTransformerUC} and filtering out biased contents in the target corpus to mitigate potential risks and harms.

\section{Proof of Theorem~\ref{thm:gmm}}
\label{sec:proof}
\begin{proof}
The MLM objective of BERT trains contextualized word embeddings to predict the masked tokens in a sequence. Formally, given an input sequence $\bs{d} = [w_1, w_2, \dots, w_n]$, a random subset of tokens (\eg, usually $15\%$ from the original sequence) $\mathcal{M}$ is selected and replaced with $\texttt{[MASK]}$ symbols. Then the BERT encoder maps the masked sequence $\hat{\bs{d}}$ to a sequence of contextualized representations $[\bs{h}_1, \bs{h}_2, \dots, \bs{h}_n]$ where $\bs{h}_i \in \mathbb{R}^{r}$ ($r = 768$ in the BERT base model). BERT is trained by maximizing the log-probability of correctly predicting every masked word with a Softmax layer over the vocabulary $V$:
\begin{equation}
\label{eq:mlm}
\max_{\bs{e},\, \bs{h},\, \bs{b}} \quad \sum_{w_i \in \mathcal{M}} \log  \frac{\exp \left(\bs{e}_{w_i}^\top \bs{h}_i + b_{w_i} \right)}{\sum_{j=1}^{|V|} \exp \left(\bs{e}_{w_j}^\top \bs{h}_i + b_{w_j} \right)},
\end{equation}
where $\bs{e}_{w_i} \in \mathbb{R}^{r}$ is the token embedding; and $b_{w_i} \in \mathbb{R}$ is a bias value for token $w_i$.

Next, we construct a multivariate GMM parameterized by the learned token embeddings $\bs{e}$ and bias vector $\bs{b}$ of BERT, and we show that the MLM objective (Eq.~\eqref{eq:mlm}) optimizes the posterior probability of contextualized embeddings $\bs{h}$ generated from this GMM. We consider the following GMM with $|V|$ mixture components, where each component $i$ is a multivariate Gaussian distribution $\mathcal{N}(\bs{\mu}_i, \bs{\Sigma}_i)$ with mean vector $\bs{\mu}_i \in \mathbb{R}^{r}$, covariance matrix $\bs{\Sigma}_i \in \mathbb{R}^{r\times r}$ and mixture weight $\pi_i$ (\ie, the prior probability) defined as follows:
$$
\bs{\mu}_i \coloneqq \bs{\Sigma}\, \bs{e}_{w_i}, \,\,
\bs{\Sigma}_i \coloneqq \bs{\Sigma}, \,\,
\pi_i \coloneqq \frac{ \exp\left(\frac{1}{2}\bs{e}_{w_i}^\top \bs{\Sigma}\, \bs{e}_{w_i} + b_{w_i} \right)}{\sum_{1\le j \le |V|} \exp\left(\frac{1}{2} \bs{e}_{w_j}^\top \bs{\Sigma}\, \bs{e}_{w_j} + b_{w_j} \right)},
$$
where all components share the same covariance matrix $\bs{\Sigma}$. 

The contextualized embeddings $\bs{h}_i$ are generated by first sampling a token $w_i$ according to the prior distribution, and then sampling from the Gaussian distribution corresponding to $w_i$, as follows:
$$
w_i \sim \text{Categorical}(\bs{\pi}), \, \bs{h}_i \sim \mathcal{N}\left(\bs{\Sigma}\, \bs{e}_{w_i},\, \bs{\Sigma}\right).
$$

Based on the above generative process, the prior probability of token $w_i$ is
$$
p(w_i) = \pi_i = \frac{ \exp\left(\frac{1}{2}\bs{e}_{w_i}^\top \bs{\Sigma}\, \bs{e}_{w_i} + b_{w_i} \right)}{\sum_{j=1}^{|V|} \exp\left(\frac{1}{2} \bs{e}_{w_j}^\top \bs{\Sigma}\, \bs{e}_{w_j} + b_{w_j} \right)},
$$
and the likelihood of generating $\bs{h}_i$ given $w_i$ is
$$
p\left(\bs{h}_i | w_i\right) = \frac{\exp\left( -\frac{1}{2} (\bs{h}_i - \bs{\Sigma}\,\bs{e}_{w_i})^\top \bs{\Sigma}^{-1} \left(\bs{h}_i - \bs{\Sigma}\,\bs{e}_{w_i}\right) \right)} {(2\pi)^{r/2} |\bs{\Sigma}|^{1/2}}.
$$
The posterior probability can be obtained using the Bayes rule:
$$
p(w_i | \bs{h}_i) = \frac{p\left(\bs{h}_i | w_i\right) p(w_i)}{\sum_{j=1}^{|V|} p\left(\bs{h}_i | w_j\right) p(w_j)},
$$
where the numerator $p\left(\bs{h}_i | w_i\right) p(w_i)$ is
$$
\frac{\exp \left(-\frac{1}{2} \bs{h}_i^\top \bs{\Sigma}^{-1} \bs{h}_i + \bs{h}_i^\top \bs{e}_{w_i} - \cancel{\frac{1}{2} \bs{e}_{w_i}^\top \bs{\Sigma}\, \bs{e}_{w_i}} + \cancel{\frac{1}{2} \bs{e}_{w_i}^\top \bs{\Sigma}\, \bs{e}_{w_i}} + b_{w_i} \right)} {(2\pi)^{r/2} |\bs{\Sigma}|^{1/2} \sum_{j=1}^{|V|} \exp\left(\frac{1}{2} \bs{e}_{w_j}^\top \bs{\Sigma}\, \bs{e}_{w_j} + b_{w_j} \right)}.
$$
The terms in the denominator are in a similar form and many common factors between the numerator and the denominator cancel out. Finally, the above posterior probability is simplified as:
$$
p(w_i | \bs{h}_i) = \frac{\exp \left(\bs{e}_{w_i}^\top \bs{h}_i + b_{w_i} \right)}{\sum_{j=1}^{|V|} \exp \left(\bs{e}_{w_j}^\top \bs{h}_i + b_{w_j} \right)},
$$
which is precisely the probability maximized by the MLM objective (Eq.~\eqref{eq:mlm}).
Therefore, the MLM pretraining objective of BERT assumes that the contextualized representations are generated from a $|V|$-component GMM.
\end{proof}

\section{Implementation Details and Parameters}\label{sec:hyperparameter}
\begin{table}[b]
\centering
\caption{Dataset statistics.}
\vspace*{-1em}
\label{tab:dataset_stats}
\scalebox{1.0}{
\begin{tabular}{cccc}
\toprule
Corpus & \# documents & \# words/doc. & Vocabulary  \\
\midrule
\textbf{NYT} & 31,997 & 690 & 25,903 \\
\textbf{Yelp} & 29,280 & 114 & 11,419 \\
\bottomrule
\end{tabular}
}
\vspace*{-1em}
\end{table}
We preprocess the corpora by discarding infrequent words that appear less than $5$ times. 
We use the default hyperparameters of baseline methods. The hyperparameters of \model are set as follows: Latent space dimension $r'=100$; 
training epochs $E=20$;
clustering loss weight $\lambda=0.1$;
DNN hidden dimensions are $500$-$500$-$1000$ for learning $f$ and $1000$-$500$-$500$ for learning $g$; the shared concentration parameter of topic vMF distributions $\kappa=10$.
We use the BERT~\cite{devlin2019bert} base model to obtain pretrained embeddings, and use Adam~\cite{kingma2015adam} with $5e-4$ learning rate to optimize the DNNs with batch size $32$.
When computing the generic document as an average of word embeddings in Eq.~\eqref{eq:doc_recons}, we only use the words that are nouns, verbs, or adjectives because they are usually the topic-indicative ones.

\end{document}